\documentclass[11pt,a4paper]{article}
\usepackage[hyperref]{acl2020}
\usepackage{times}
\usepackage{latexsym}
\usepackage{url}
\usepackage{paralist}
\usepackage{multirow}
\usepackage{verbatim}
\usepackage{amsfonts}
\usepackage{amssymb}
\usepackage{amsmath}
\usepackage{algorithm}
\usepackage{algpseudocode}
\usepackage{graphics}
\usepackage{epsfig}
\usepackage{float}
\usepackage{multicol}
\usepackage{makecell}

\usepackage{enumitem}
\usepackage{amsmath}

\usepackage{microtype}

\aclfinalcopy 
\def\aclpaperid{2623} 

\title{Learning Spoken Language Representations with\\ Neural Lattice Language Modeling}

\author{Chao-Wei Huang \qquad Yun-Nung (Vivian) Chen \\
  Department of Computer Science and Information Engineering \\
  National Taiwan University, Taipei, Taiwan \\
  \texttt{f07922069@csie.ntu.edu.tw \quad y.v.chen@ieee.org}}

\date{}

\begin{document}
\maketitle
\begin{abstract}
Pre-trained language models have achieved huge improvement on many NLP tasks.
However, these methods are usually designed for written text, so they do not consider the properties of spoken language. 
Therefore, this paper aims at generalizing the idea of language model pre-training to lattices generated by recognition systems.
We propose a framework that trains neural lattice language models to provide contextualized representations for spoken language understanding tasks.
The proposed two-stage pre-training approach reduces the demands of speech data and has better efficiency.
Experiments on intent detection and dialogue act recognition datasets demonstrate that our proposed method consistently outperforms strong baselines when evaluated on spoken inputs.\footnote{The scource code is available at: \url{https://github.com/MiuLab/Lattice-ELMo}.}
\end{abstract}

\section{Introduction}
The task of spoken language understanding (SLU) aims at extracting useful information from spoken utterances.
Typically, SLU can be decomposed with a two-stage method:
1) an accurate automatic speech recognition (ASR) system transcribes the input speech into texts, and then
2) language understanding techniques are applied to the transcribed texts.
These two modules can be developed separately, so most prior work developed the backend language understanding systems based on manual transcripts~\cite{yao2014spoken,guo2014joint,mesnil2014using,goo-etal-2018-slot}.

\begin{figure}[t!]
\centering 
\includegraphics[width=0.9\linewidth]{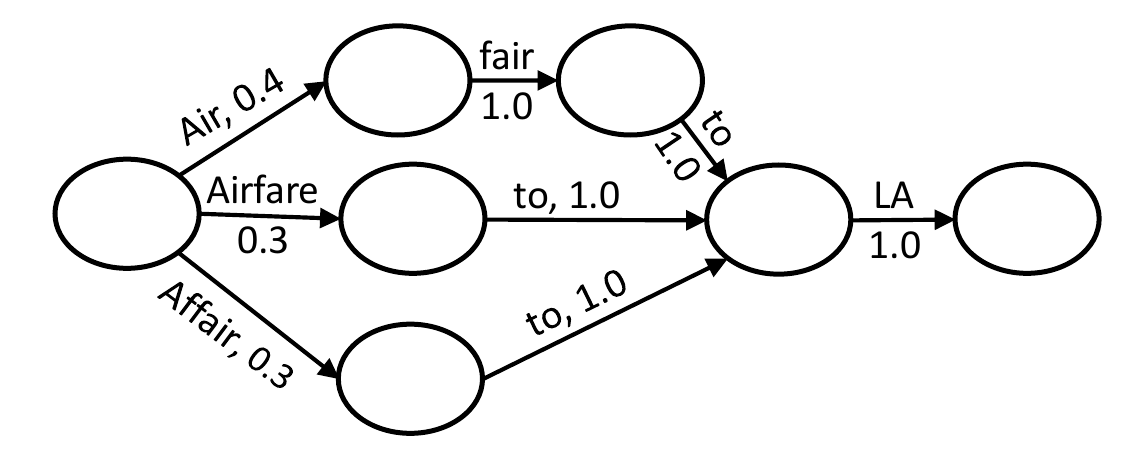}
 \caption{Illustration of a lattice.} 
\label{fig:lattice} 
\vspace{-3mm}
\end{figure}

Despite the simplicity of the two-stage method, prior work showed that a tighter integration between two components can lead to better performance.
Researchers have extended the ASR 1-best results to n-best lists or word confusion networks in order to preserve the ambiguity of the transcripts. ~\cite{tur2002improving,hakkani2006beyond,henderson2012discriminative,Tr2013SemanticPU,masumura2018neural}.
Another line of research focused on using lattices produced by ASR systems.
Lattices are directed acyclic graphs (DAGs) that represent multiple recognition hypotheses.
An example of ASR lattice is shown in Figure~\ref{fig:lattice}. 
\citet{ladhak2016latticernn} introduced LatticeRNN, a variant of recurrent neural networks (RNNs) that generalize RNNs to lattice-structured inputs in order to improve SLU.
\citet{zhang-yang-2018-chinese} proposed a similar idea for Chinese name entity recognition.
\citet{sperber-etal-2019-self,xiao-etal-2019-lattice,zhang-etal-2019-lattice} proposed extensions to enable the transformer model~\cite{vaswani2017attention} to consume lattice inputs for machine translation.
\citet{huang2019adapting} proposed to adapt the transformer model originally pre-trained on written texts to consume lattices in order to improve SLU performance.
\citet{buckman-neubig-2018-neural} also found that utilizing lattices that represent multiple granularities of sentences can improve language modeling.

With recent introduction of large pre-trained language models (LMs) such as ELMo~\cite{peters-etal-2018-deep}, GPT~\cite{Radford2018ImprovingLU} and BERT~\cite{devlin-etal-2019-bert}, we have observed huge improvements on natural language understanding tasks.
These models are pre-trained on large amount of written texts so that they provide the downstream tasks with high-quality representations.
However, applying these models to the spoken scenarios poses several discrepancies between the pre-training task and the target task, such as the domain mismatch between written texts and spoken utterances with ASR errors.
It has been shown that fine-tuning the pre-trained language models on the data from the target tasks can mitigate the domain mismatch problem~\cite{howard-ruder-2018-universal,chronopoulou-etal-2019-embarrassingly}.
\citet{siddhant2018unsupervised} focused on pre-training a language model specifically for spoken content with huge amount of automatic transcripts, which requires a large collection of in-domain speech.

In this paper, we propose a novel spoken language representation learning framework, which focuses on learning contextualized representations of lattices based on our proposed lattice language modeling objective.
The proposed framework consists of two stages of LM pre-training to reduce the demands for lattice data.
We conduct experiments on benchmark datasets for spoken language understanding, including intent classification and dialogue act recognition.
The proposed method consistently achieves superior performance, with relative error reduction ranging from 3\% to 42\% compare to pre-trained sequential LM.

\begin{figure*}[t!]
\centering 
\includegraphics[width=\linewidth]{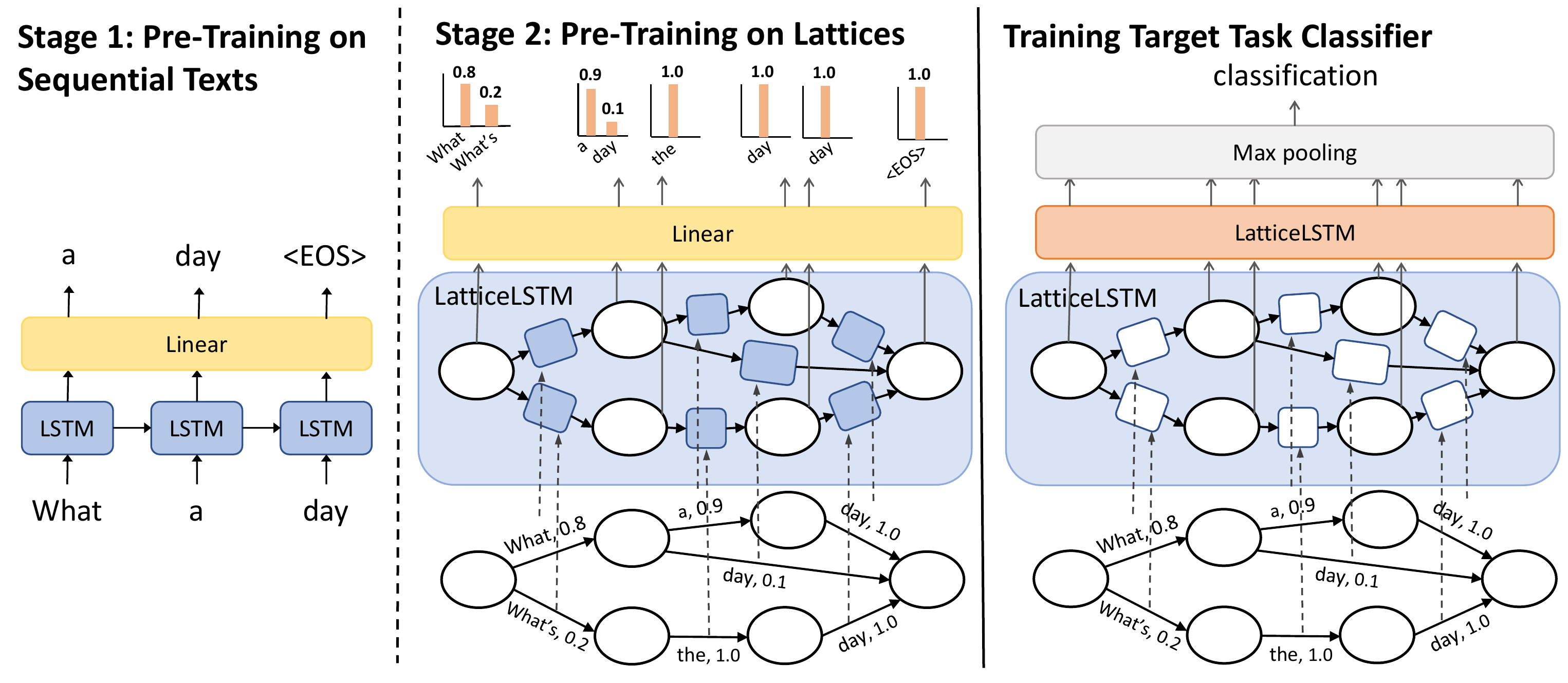}
 \caption{Illustration of the proposed framework. The weights of the pre-trained LatticeLSTM LM are fixed when training the target task classifier (shown in white blocks), while the weights of the newly added LatticeLSTM classifier are trained from scratch (shown in colored block).} 
\label{fig:framework} 
\end{figure*}

\section{Neural Lattice Language Model}
The two-stage framework that learns contextualized representations for spoken language is proposed and detailed below.

\subsection{Problem Formulation}
In the SLU task, the model input is an utterance $X$ containing a sequence of words $X = [x_1, x_2, \cdots , x_{|X|}]$, and the goal is to map $X$ to its corresponding class $y$.
The inputs can also be stored in a lattice form, where we use edge-labeled lattices in this work.
A lattice $L = \{N, E \}$ is defined by a set of $|N|$ nodes $N = \{n_1, n_2, \cdots, n_{|N|} \}$ and a set of $|E|$ transitions $E = \{e_1, e_2, \cdots, e_{|E|} \}$.
A weighted transition is  defined as $e = \{prev[e], next[e], w[e], P(e)\}$, where $prev[e]$ and $next[e]$ denote the previous node and next node respectively, $w[e]$ denotes the associated word, and $P(e)$ denotes the transition probability.
We use $in[n]$ and $out[n]$ to denote the sets of incoming and outgoing transitions of a node $n$.
$L_{<n} = \{N_{<n}, E_{<n}\}$ denotes the sub-lattice which consists of all paths between the starting node and a node $n$.

\subsection{LatticeRNN}
The LatticeRNN~\cite{ladhak2016latticernn} model generalizes sequential RNN to lattice-structured inputs.
It traverses the nodes and transitions of a lattice in a topological order.
For each transition $e$, LatticeRNN takes $w[e]$ as input and the representation of its previous node $h[prev[e]]$ as the previous hidden state, and then produces a new hidden state of $e$, $h[e]$.
The representation of a node $h[n]$ is obtained by pooling the hidden states of the incoming transitions.
In this work, we employ the \textit{WeightedPool} variant proposed by~\citet{ladhak2016latticernn}, which computes the node representation as
\begin{equation*}
    h[n] = \sum_{e \in in[n]} P(e) \cdot h[e].
\end{equation*}

Note that we can represent any sequential text as a linear-chain lattice, so LatticeRNN can be seen as a strict generalization of RNNs to DAG-like structures.
This property enables us to initialize the weights in a LatticeRNN with the weights of a RNN as long as they use the same recurrent cell.

\subsection{Lattice Language Modeling}
Language models usually estimate $p(X)$ by factorizing it into
\begin{equation*}
    p(X) = \prod_{t=0}^{|X|} p(x_t \mid X_{<t}),
\end{equation*}
where $X_{<t} = [x_1, \cdots, x_{t-1}]$ denotes the previous context.
Training a LM is essentially asking the model to predict a distribution of the next word given the previous words.
We extend the sequential LM analogously to \emph{lattice language modeling}, where the model is expected to predict the next transitions of a node $n$ given $L_{<n}$.
The ground truth distribution is therefore defined as:
\begin{align*}
    & p(w \mid L_{<n}) \\
    & =
    \begin{cases}
        P(e), & \text{if } \exists e \in out[n] \text{ s.t. } w[e] = w \\
        0,    & \text{otherwise.}
    \end{cases}
\end{align*}

LatticeRNN is adopted as the backbone of our lattice language model.
Since the node representation $h[n]$ encodes all information of $L_{<n}$, we pass $h[n]$ to a linear decoder to obtain the distribution of next transitions:
\begin{equation*}
    p_{\theta}(w \mid h[n]) = \texttt{softmax}(W^T h[n]),
\end{equation*}
where $\theta$ denotes the parameters of the LatticeRNN and $W$ denotes the trainable parameters of the decoder.
We train our lattice language model by minimizing the KL divergence between the ground truth distribution $p(w \mid L_{<n})$ and the predicted distribution $p_{\theta}(w \mid h[n])$.

Note that the objective for training sequential LM is a special case of the lattice language modeling objective defined above, where the inputs are linear-chain lattices.
Hence, a sequential LM can be viewed as a lattice LM trained on linear-chain lattices only.
This property inspires us to pre-train our lattice LM in a 2-stage fashion described below.

\subsection{Two-Stage Pre-Training}
Inspired by ULMFiT~\cite{howard-ruder-2018-universal}, we propose a two-stage pre-training method to train our lattice language model.
The proposed method is illustrated in Figure~\ref{fig:framework}.

\begin{itemize}
    \item Stage 1: Pre-train on sequential texts\\
    In the first stage, we follow the recent trend of pre-trained LMs by pre-training a bidirectional LSTM~\cite{hochreiter1997long} LM on general domain text corpus.
    Here the cell architecture is the same as ELMo~\cite{peters-etal-2018-deep}.
    \item Stage 2: Pre-train on lattices\\
In this stage, we use a bidirectional LatticeLSTM with the same cell architecture as the LSTM pre-trained in the previous stage. Note that in the backward direction we use reversed lattices as input.
We initialize the weights of the LatticeLSTM with the weights of the pre-trained LSTM.
The LatticeLSTM is further pre-trained on lattices from the training set of the target task with the lattice language modeling objective described above.
\end{itemize}

We consider this two-stage method more \emph{approachable} and \emph{efficient} than directly pre-training a lattice LM on large amount of lattices because
1) general domain written data is much easier to collect than lattices which require spoken data, and
2) LatticeRNNs are considered less efficient than RNNs due to the difficulty of parallelization in computing.

\subsection{Target Task Classifier Training}
After pre-training, our model is capable of providing representations for lattices.
Following~\cite{peters-etal-2018-deep}, the pre-trained lattice LM is used to produce contextualized node embeddings for downstream classification tasks, as illustrated in the right part of Figure~\ref{fig:framework}.
We use the same strategy as~\citet{peters-etal-2018-deep} to linearly combine the hidden states from different layers into a representation for each node.
The classifier is a newly added 2-layer LatticeLSTM, which takes the node representations as input, followed by max-pooling over nodes, a linear layer and finally a softmax layer.
We use the cross entropy loss to train the classifier on each target classification tasks.
Note that the parameters of the pre-trained lattice LM are fixed during this stage.

\begin{table*}[t!]
\centering
\begin{tabular}{|c|c|l|cc|cc|}
\hline
    \multicolumn{3}{|c|}{}  & \bf ATIS & \bf SNIPS & \bf SWDA & \bf MRDA \\
\hline\hline
\multirow{2}{*}{Manual} & (a) & biLSTM        &   -   &  97.00     &  71.19    &  79.99    \\
\cline{2-7}
& (b) & (a) + ELMo    &  -    & 96.80      &    72.18  & 81.48      \\
\hline
\multirow{2}{*}{Lattice oracle} & (c) & biLSTM        &   92.97   &  94.02     &  63.92    &  70.49   \\
\cline{2-7}
& (d) & (c) + ELMo    &  96.21    & 95.14      &    65.14  & 73.34      \\
\hline\hline
\multirow{3}{*}{ASR 1-Best} & (e) & biLSTM        &  91.60    & 91.89      & 60.54      & 67.35     \\
\cline{2-7}
& (f) & (e) + ELMo    & 94.99     & 91.98      &    61.65  &    68.52  \\
\cline{2-7}
& (g) & BERT-base    & \bf 95.97     & 93.29      &    61.23  &    67.90  \\
\hline
\multirow{5}{*}{Lattices} & (h) & biLatticeLSTM \qquad \qquad &   91.69   & 93.43      & 61.29     & 69.95     \\
\cline{2-7}
& (i) & Proposed  & 95.84     & \bf 95.37      & \bf 62.88  & \bf 72.04  \\
\cline{2-7}
& (j) & (i) w/o Stage 1  & 94.65     & 95.19      & 61.81     &  71.71    \\
\cline{2-7}
& (k) & (i) w/o Stage 2  & 95.35     & 94.58      & 62.41     &  71.66    \\
\cline{2-7}
& (l) & (i) evaluated on 1-best  & 95.05     & 92.40      & 61.12     &  68.04    \\
\hline
\end{tabular}
\renewcommand\thetable{2}
\caption{Results of our experiments in terms of accuracy (\%). Some audio files in ATIS are missing, so the testing sets of manual transcripts and ASR transcripts are different. Hence, we do not report the results for ATIS using manual transcripts. The best results obtained by using ASR output for each dataset are marked in bold.}
\label{tab:result}
\end{table*}

\begin{table}[H]
\centering
\small
\begin{tabular}{|l|rr|rr|}
\hline
                 & \bf ATIS & \bf SNIPS & \bf SWDA & \bf MRDA \\
\hline\hline
\textbf{Train}             & 4,478 & 13,084       & 103,326       & 73,588       \\
\hline
\textbf{Valid}             & 500 & 700       & 8,989        & 15,037        \\
\hline
\textbf{Test}              & 869 & 700       & 15,927        & 14,800        \\
\hline
\textbf{\#Classes}            & 22 &  7      &   43      & 5   \\
\hline
\textbf{WER(\%)}               & 15.55 & 45.61      & 28.41        & 32.04   \\
\hline
\textbf{Oracle WER}               & 9.19 & 18.79      & 17.15        & 21.53   \\
\hline
\end{tabular}
\vspace{-1mm}
\renewcommand\thetable{1}
\caption{Data statistics.}
\label{tab:dataset}
\end{table}

\section{Experiments}
In order to evaluate the quality of the pre-trained lattice LM, we conduct the experiments for two common tasks in spoken language understanding.

\subsection{Tasks and Datasets}
Intent detection and dialogue act recognition are two common tasks about spoken language understanding. 
The benchmark datasets used for intent detection are ATIS (Airline Travel Information Systems)~\cite{hemphill1990atis,dahl1994expanding,tur2010left} and SNIPS~\cite{coucke2018snips}.
We use the NXT-format of the Switchboard~\cite{stolcke-etal-2000-dialogue} Dialogue Act Corpus~(SWDA)~\cite{calhoun2010nxt} and the ICSI Meeting Recorder Dialogue Act Corpus~(MRDA)~\cite{shriberg-etal-2004-icsi} for benchmarking dialogue act recognition.
The SNIPS corpus only contains written text, so we synthesize a spoken version of the dataset using a commercial text-to-speech service.
We use an ASR system trained on WSJ~\cite{Paul:1992:DWS:1075527.1075614} with Kaldi~\cite{povey2011kaldi} to transcribe ATIS, and an ASR system released by Kaldi to transcribe other datasets.
The statistics of datasets are summarized in Table~\ref{tab:dataset}.
All tasks are evaluated with overall classification accuracy.

\subsection{Model and Training Details}
In order to conduct fair comparison with ELMo~\cite{peters-etal-2018-deep}, we directly adopt their pre-trained model as our pre-trained sequential LM.
The hidden size of the LatticeLSTM classifier is set to 300.
We use \texttt{adam} as the optimizer with learning rate 0.0001 for LM pre-training and 0.001 for training the classifier.
The checkpoint with the best validation accuracy is used for evaluation.

\subsection{Results}
The results in terms of the classification accuracy are shown in Table~\ref{tab:result}.
All reported numbers are averaged over at least three training runs.
Rows (a) and (b) can be considered as the performance upperbound, where we use manual transcripts to train and evaluate the models.
We also use BERT-base~\cite{devlin-etal-2019-bert} as a strong baseline, which takes ASR 1-best as the input (row (g)).
Compare with the results on manual transcripts, using ASR results largely degrades the performance due to recognition errors, as shown in rows (e)-(g).
In addition, adding pre-trained ELMo embeddings brings consistent improvement over the biLSTM baseline, except for SNIPS when using manual transcripts (row (b)).
The baseline models trained on ASR 1-best are also evaluated on lattice oracle paths. We report the results as the performance upperbound for the baseline models (rows (c)-(d)).


In the lattice setting, the baseline bidirectional LatticeLSTM~\cite{ladhak2016latticernn} (row (h)) consistently outperforms the biLSTM with 1-best input (row (e)), demonstrating the importance of taking lattices into account.
Our proposed method achieves the best results on all datasets except for ATIS (row(i)), with relative error reduction ranging from 3.2\% to 42\% compare to biLSTM+ELMo (row(f)).
The proposed method also achieves performance comparable to BERT-base on ATIS.
We perform ablation study for the proposed two-stage pre-training method and report the results in rows (j) and (k).
It is clear that skipping either stage degrades the performance on all datasets, demonstrating that both stages are crucial in the proposed framework.
We also evaluate the proposed model on 1-best results (row (l)). The results show that it is still beneficial to use lattice as input after fine-tuning.

\section{Conclusion}
In this paper, we propose a spoken language representation learning framework that learns contextualized representation of lattices.
We introduce the lattice language modeling objective and a two-stage pre-training method that efficiently trains a neural lattice language model to provide the downstream tasks with contextualized lattice representations.
The experiments show that our proposed framework is capable of providing high-quality representations of lattices, yielding consistent improvement on SLU tasks.

\section*{Acknowledgement}
We thank reviewers for their insightful comments. This work was financially supported from the Young Scholar Fellowship Program by Ministry of Science and Technology (MOST) in Taiwan, under Grant 109-2636-E-002-026.

\bibliography{acl2020}
\bibliographystyle{acl_natbib}

\end{document}